\documentclass[sigconf, nonacm]{acmart}
\AtBeginDocument{%
  }

\usepackage{textgreek}
\usepackage{algorithm}
\usepackage[noend]{algpseudocode} 
\usepackage{amsmath}
\usepackage[x11names]{xcolor}
\usepackage{enumitem}
\usepackage{appendix} 

\usepackage{times}
\usepackage{latexsym}

\usepackage[T1]{fontenc}

\usepackage[utf8]{inputenc}

\usepackage{microtype}

\usepackage{graphicx}
\usepackage{booktabs}
\usepackage{multirow}
\usepackage[table]{xcolor}
\definecolor{mygr}{RGB}{0,128,0}
\usepackage{xparse}
\newcommand{\COMMENTLLAMA}[1]{{\color{mygr} $\triangleright$ {#1}}}
\definecolor{EI}{RGB}{108,144,213}
\definecolor{SN}{RGB}{118,155,93}
\definecolor{TF}{RGB}{234,164,86}
\definecolor{PJ}{RGB}{235,116,116}
\usepackage{tabularx}
\usepackage{amsthm}
\usepackage{soul}
\begin{document}


\title{Multi-Personality Generation of LLMs at Decoding-time}

\author{Rongxin Chen}
\authornotemark[2]
\affiliation{%
  \institution{Institute of Computing Technology,
Chinese Academy of Sciences\\ University of Chinese Academy of Sciences}
  \city{Beijing}
  \country{China}}
\email{chenrongxin24s@ict.ac.cn}

\author{YunFan Li}
\authornotemark[2]
\affiliation{%
  \institution{Institute of Computing Technology,
Chinese Academy of Sciences\\ University of Chinese Academy of Sciences}
  \city{Beijing}
  \country{China}}
\email{liyunfan24s@ict.ac.cn}

\author{Yige Yuan}
\authornotemark[2]
\affiliation{%
  \institution{Institute of Computing Technology,
Chinese Academy of Sciences\\ University of Chinese Academy of Sciences}
  \city{Beijing}
  \country{China}}
\email{yuanyige20z@ict.ac.cn}

\author{Bingbing Xu}
\authornote{Corresponding author.}
\authornotemark[2]
\affiliation{%
  \institution{Institute of Computing Technology, Chinese Academy of Sciences}
  \city{Beijing}
  \country{China}}
\email{xubingbing@ict.ac.cn}

\author{Huawei Shen}
\authornote{State Key Laboratory of AI Safety of Chinese Academy of Sciences (CAS).}
\affiliation{%
  \institution{Institute of Computing Technology, Chinese Academy of Sciences}
  \city{Beijing}
  \country{China}}
\email{shenhuawei@ict.ac.cn}


\begin{abstract}

Multi-personality generation for LLMs, enabling simultaneous embodiment of multiple personalization attributes, is a fundamental challenge. Existing retraining-based approaches are costly and poorly scalable, while decoding-time methods often rely on external models or heuristics, limiting flexibility and robustness.
In this paper, we propose a novel Multi-Personality Generation (MPG) framework under the decoding-time combination paradigm. It flexibly controls multi-personality without relying on scarce multi-dimensional models or extra training, leveraging implicit density ratios in single-dimensional models as a "free lunch" to reformulate the task as sampling from a target strategy aggregating these ratios.
To implement MPG efficiently, we design Speculative Chunk-level based Rejection sampling (SCR), which generates responses in chunks and parallelly validates them via estimated thresholds within a sliding window. This significantly reduces computational overhead while maintaining high-quality generation.
Experiments on MBTI personality and Role-Playing demonstrate the effectiveness of MPG, showing improvements up to 16\%–18\%. Code and data are available at \url{https://github.com/Libra117/MPG}.

\end{abstract}

\maketitle

\section{Introduction}

\begin{figure}[t]
  \centering
  \includegraphics[width=0.9\linewidth, height=6.5cm]{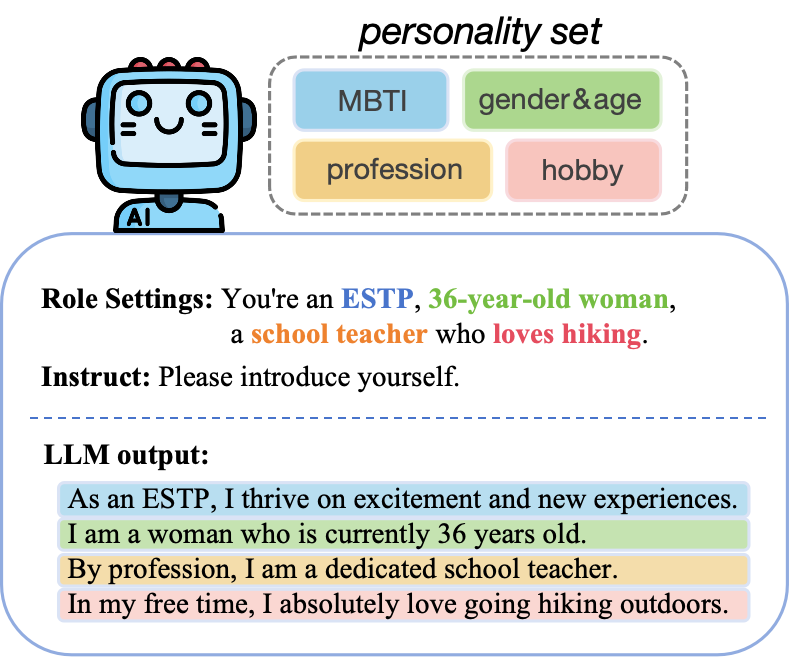}
  \caption{Role-Playing in Multi-personality generation.}
  \label{fig.1}
\end{figure}

In recent years, Large Language Models (LLMs)~\cite{Zhao2023ASO} have witnessed rapid development, with parameter scales surging from billions to trillions and training data covering diverse domains such as text and code~\cite{Chen2021EvaluatingLL,Li2024PersonalLA}, thereby endowing them with capabilities in reasoning and semantic understanding~\cite{Huang2022TowardsRI}. Within this context, multi-personality generation has emerged as a critical task, which requires LLMs to transform from general-purpose tools into personalized intelligences capable of understanding and addressing the nuanced needs of individual users, and is widely applied in scenarios such as personalized question answering, artificial assistants, and intelligent customer service~\cite{Li2024PersonalLA,Guo2024LargeLM,Cheng2024ExploringLL,NEURIPS2024_1cb57fcf}.

Multi-personality generation requires generated text to simultaneously embody multiple, potentially interacting personalization attributes. Figure~\ref{fig.1} illustrates multi-personality generation in role-play, where the personality set specifies multi-dimensional requirements, and LLMs need to take all these dimensions into account when generating responses.
However, the inherent complexity of modeling multiple personalizations concurrently, combined with the challenge of balancing and integrating these traits within general-purpose LLMs, amplifies the difficulty of this research.

Existing studies can be broadly categorized into retraining-based and decoding-time methods. 
Retraining methods seek to encode multiple preference dimensions into a single model during training via multi-objective optimization. This involves techniques like multi-objective reinforcement learning where models learn to balance competing preferences through iterative feedback and weighted-loss supervised fine-tuning, which assigns differential weights to preference signals to guide parameter updates~\cite{Harland2024AdaptiveAD,Zhou2023BeyondOA}.
But these approaches suffer from notable drawbacks: they incur high training costs and exhibit poor scalability when adapting to new preferences, often requiring full retraining.
In contrast, decoding-time methods~\cite{Huang2024DeALDA,Khanov2024ARGSAA,NEURIPS2024_3950f6bf} circumvent the need for retraining by guiding the decoding process through external tools. 
Specifically, they require a model capable of evaluating the quality of responses at the multi-personality level, and leverage this model as reward model ~\cite{Chen2024PADPA,Yang2024RewardsinContextMA,Khanov2024ARGSAA} or aligner ~\cite{Yang2024MetaalignerTG} to shape output, or introduce target preference signals directly into generation via prompt learning \cite{Chen2024PADPA,Ding2021OpenPromptAO}. 
Yet, they depend heavily on external models that are hard to acquire and struggle to handle preferences that change dynamically.

To provide efficiency and practicality in dynamic scenarios where user preferences evolve, combining multiple models at decode time rather than relying on external reward models has emerged as a trend.
Specifically, such approaches require the existence of a model for each dimension (e.g., the model for Hobby is a single-dimensional model in Figure~\ref{fig.1}). 
And they involve either combining model parameters across multi-dimensions ~\cite{Jang2023PersonalizedSP,Ram2023RewardedST,Lu2023InferringPF} or linearly combining the prediction logits of models from multiple dimensions to generate the next token ~\cite{Shi2024DecodingTimeLM}. However, their combinatorial mechanisms remain heuristic, with performance constrained by the capabilities of individual constituent models and lacking robustness.
Overall, there remains a lack of a model capable of flexibly integrating diverse personality dimensions while maintaining efficiency, scalability, and robust performance across dynamic scenarios.

In this paper, we propose a novel \textbf{M}ulti-\textbf{P}ersonality \textbf{G}eneration (MPG) framework under the decoding-time combination paradigm. It flexibly controls multi-personality traits in generation, ensuring robustness without relying on scarce multi-personality models or extra training.
Specifically, the core lies in how to leverage these single-dimensional models to induce better responses. Single-dimensional models are usually obtained through alignment (e.g., DPO, PPO) on the original model. We find that the implicit density ratio encoded during the alignment process captures the specific preference patterns of each model relative to the reference model at no extra cost as a "free lunch". This perspective enables us to reformulate multi-personality generation as sampling from a target strategy that aggregates these ratios.
From this perspective, the probability of sampling a response from the target strategy is proportional to the weighted sum of its density ratios in single-dimensional models relative to the base model.
This insight is naturally reflected in the acceptance probability of rejection sampling, making it possible to induce the ideal response of the target strategy by performing rejection sampling on the responses generated by the base policy through this probability.

To implement the above framework, we first considered traditional rejection sampling: generating multiple responses from a base policy and fusing model scores to calculate acceptance thresholds. However, this is time-consuming with low response utilization, as most responses only serve threshold calculation.
To address this, we propose 
\textbf{S}peculative \textbf{C}hunk-level based \textbf{R}ejection sampling
(SCR), which generates responses in chunks, parallelly validating token chunks across single-dimensional models, and dynamically estimating acceptance thresholds via the maxima within the chunk sliding window, avoiding repeated full response sampling from the base policy.
Extensive experiments in MBTI personality simulation and Role-Playing scenarios validate its effectiveness and efficiency, with up to 16\%–18\% improvement over baselines. Data and code are open-sourced.

Overall, our contributions can be summarized as follows:
\begin{itemize}
\item
\textbf{Promising framework}: we propose a novel MPG framework under the decoding-time combination paradigm, enabling flexible control of multi-personality without extra training, by leveraging implicit density ratios from single-dimensional models as a free lunch.

\item \textbf{Practical method}: we design SCR, a novel decoding-time algorithm that performs speculative chunk-level rejection sampling with parallel multi-preference scoring, substantially reducing the computational overhead of multiple large-model preference evaluations during decoding.

\item \textbf{Validated effectiveness}: we validate the framework and method in MBTI simulation and role-play scenarios, achieving 16\%–18\% improvement over baselines, with open-sourced data and code.

\end{itemize}

\begin{figure*}[ht]
    \centering
    \includegraphics[width=0.85\linewidth, height = 8.5cm]{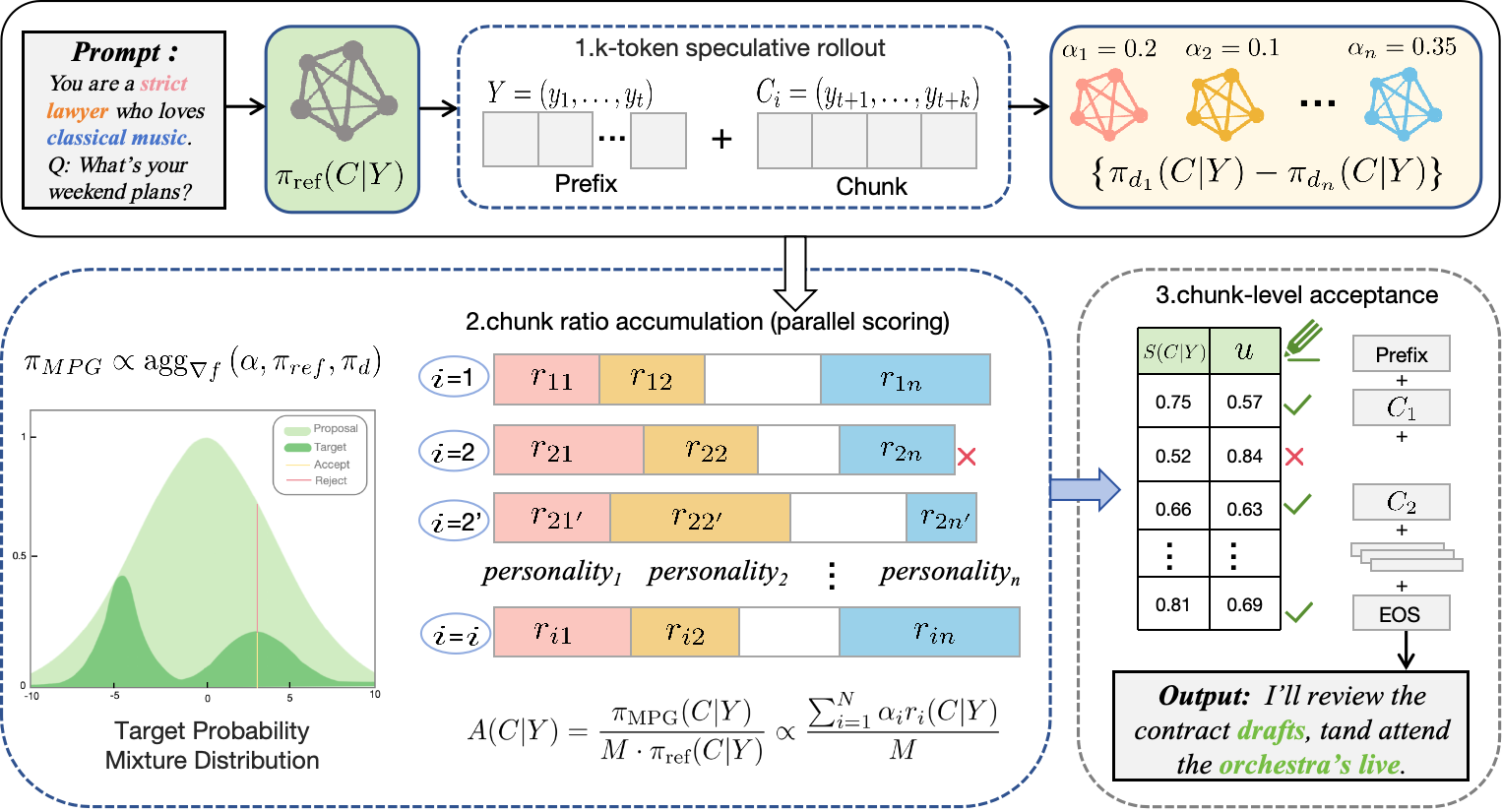}
    \caption{An illustration of the proposed Speculative Chunk-level based Rejection sampling(SCR) algorithm. 
Given a prompt, the reference model generates $k$-token speculative chunks, which are scored by multiple preference models in parallel via weighted density ratios relative to the reference. 
Chunk-level acceptance is performed using the aggregated score, with prefix salvage applied if a full chunk is rejected. 
This integrates speculative decoding with rejection sampling to efficiently sample from the multi-preference target distribution while reducing large-model evaluations.
}
    \label{fig:method}
\end{figure*}

\section{Related Work}

In the following, we introduce the related work from three aspects: RLHF,  which serves as the basis of the single-dimensional model, decoding-time alignment, and Multi-dimensional Personalization.

\textbf{Reinforcement Learning from Human Feedback.}
RLHF~\cite{Ziegler2019FineTuningLM,Ouyang2022TrainingLM,Bai2022TrainingAH,Dubois2023AlpacaFarmAS,Bai2022ConstitutionalAH} serves as the dominant paradigm for aligning Large Language Models with human preferences. It traditionally employs a three-stage pipeline: Supervised Fine-Tuning (SFT), training a reward model from human feedback, and then enhancing the policy through a reinforcement learning algorithm, most notably PPO~\cite{Schulman2017ProximalPO}. More recent approaches, such as DPO~\cite{Rafailov2023DirectPO}, offer an alternative to this final stage by directly optimizing the policy on preference data, bypassing the need for an explicit reward model and RL optimization.

\textbf{Decoding-time Alignment.}
Decoding-time alignment methods emerged as an effective approach to satisfy diverse user requirements, which can adapt to different objectives without retraining. 
PAD~\cite{Chen2024PADPA} introduces token-level personalized rewards by decoupling
text generation dynamics from user preferences, enabling dynamic
adjustment of base model predictions during inference. It
leverages a single policy and reward model to generalize
to unseen preferences.
MOD~\cite{Shi2024DecodingTimeLM} employs a closed-form solution derived via Legendre transformation to combine predictions from multiple base models-each aligned to distinct objectives-during decoding. By weighting model outputs algebraically, MOD achieves precise control over objective weights.
Drift~\citep{kim-etal-2025-drift} extends this direction by inferring user preferences through attribute-level decomposition and combining attribute signals directly at decoding time, enabling effective few-shot personalization without explicit reward models.

\textbf{Multi-dimensional Personalization.} 
For the more complex multi-dimensional personalization alignment challenge of having models satisfy multiple (or even conflicting) objectives simultaneously (e.g., balancing usefulness and harmlessness, or simulating a specific MBTI/Role-Playing persona), the research community has explored different strategies. The main approaches include training a single model to optimize a weighted multi-objective function (e.g., MORLHF, MODPO~\cite{Zhou2023BeyondOA}), combining the parameters of independently-trained single-preference models (e.g., DPO Soups~\cite{Jang2023PersonalizedSP}) during decoding, and bootstrapping by combining reward signals or model predictions at decoding (e.g., PAD~\cite{Chen2024PADPA}, MOD~\cite{Shi2024DecodingTimeLM}). Drift~\citep{kim-etal-2025-drift} decomposes user preferences into interpretable attributes and composes them at decoding time, providing a new perspective on multi-dimensional personalization.

\section{Methodology}

This section details the MPG framework by first formally defining the multi-personality generation problem, then establishing the theoretical underpinnings of our method based on the density ratio principle, and finally introducing our efficient implementation, Speculative Chunk-level based Rejection sampling (SCR).

\subsection{Problem Definition and Formulation}

Our paper addresses the problem of generating text that exhibits multiple personality attributes at decode time for LLMs. Specifically, given an input $x$ and preference weight vector $\mathbf{\alpha}=[\alpha_1,\ldots,\alpha_N]$ (with constraints $\alpha_i \geq 0$ and $\sum_{i=1}^N \alpha_i=1$),
generate a text sequence $y$ that simultaneously embodies these personality features under a set of $N$ desired personality attributes $\{d_1, \dots, d_N\}$.

Next, we formalize our problem. Given:
\begin{itemize}
\item A reference language model $\pi_{\text{ref}}(y|x)$, which represents the basic language generation ability of the model.
\item $N$ single-attribute preference models $\{\pi_{d_i}(y|x)\}_{i = 1}^N$, capturing the specific preference for the $i$-th personality attribute (relative to $\pi_{\text{ref}}(y|x)$). 
\end{itemize}

For $\pi_{d_i}(y|x)$, the RLHF~\cite{Ziegler2019FineTuningLM,Ouyang2022TrainingLM,Bai2022TrainingAH,Dubois2023AlpacaFarmAS} training objective under $f$-divergence regularization for the reward functions $\mathcal R_i$, the closed form
of $\pi_{d_i}$ can be written as:
\begin{equation}
\pi_{d_i}=\underset{\pi \in \mathcal{S}}{\operatorname{\arg max}} \underset{\substack{x \sim \mathcal{X} \\ y \sim \pi(y|x)}}{\mathbb{E}}\left[\mathcal{R}_{i}(y|x)\right]-\beta \underset{\substack{x \sim \mathcal{X} \\ y \sim \pi_{\text{ref}}(y|x)}}{\mathbb{E}} f\left(\frac{\pi(y|x)}{\pi_{\text{ref}}(y|x)}\right)
\end{equation}
where $\beta > 0$ is a regularization parameter. For the preference on dimension $i$, our single-attribute preference model $\pi_{d_i}(y|x)$ can be viewed as an optimal policy trained to maximize an implicit reward function ${\mathcal R}_i(y|x)$. As stated in the work of MOD~\cite{Shi2024DecodingTimeLM}, if the barrier function $f$ is continuously differentiable and strongly convex on $\mathbb{R}_{+}$, the following closed-form bijection between $\pi_{d_i}$ and the corresponding $\mathcal{R}_{i}$ can be obtained:

\begin{equation}
\pi_{di}(y|x) = \dfrac{1}{Z_i(x)}\pi_{\text{ref}}(y|x)  (\nabla f)^{(-1)}\left(\frac{1}{\beta}   \mathcal{R}_i(y|x)\right)
\end{equation}

\begin{equation}
\mathcal{R}_i(y\vert x)=\beta\nabla f\left(\frac{\pi_{di}(y\vert x)}{\pi_{ref}(y\vert x)}\right)+\beta Z_i(x)
\end{equation}
where $Z_i(x)$ is the normalization factor with respect to $x$, $(\nabla f)^{(-1)}$ denotes the inverse function of $\nabla f$.

Defining the ultimate objective of multiple personality decoding of a combination of reward functions
$\sum_{i=1}^N\alpha_i{\mathcal R}_i(y|x)$. 
The optimal policy of the combination reward can be derived as:

\begin{equation}
\pi^{\star}(y|x) = \dfrac{1}{Z^{\star}(x)}\pi_{\text{ref}}(y|x)  (\nabla f)^{(-1)}\left(\frac{1}{\beta} \sum_{i=1}^N \alpha_i   \mathcal{R}_i(y|x)\right)
\end{equation}

Apply $\mathcal{R}_i$ and simplify to get $\pi_\mathrm{MPG}(y|x)$:

\begin{equation}
\label{(5)}
\pi_\mathrm{MPG}(y|x) = \frac{\pi_{\text{ref}}(y|x)}{Z(x)}   \left(\nabla f\right)^{(-1)}\left(\frac{1}{\beta} \sum_{i=1}^{N} \alpha_{i}   \nabla f\left(\frac{\pi_{d_i}(y|x)}{\pi_{\text{ref}}(y|x)}\right)\right)
\end{equation}

where $Z(x)$ and $Z^\star(x)$ are normalization factors. 

Unlike approaches that directly optimize the combined reward during training time via reinforcement learning~\cite{Ziegler2019FineTuningLM,Shi2024DecodingTimeLM,Ouyang2022TrainingLM,Bai2022TrainingAH,Dubois2023AlpacaFarmAS,Kim2023AligningLL}, our method draws inspiration from the mixture of models during decoding~\cite{Wortsman2022ModelSA,Lu2023InferringPF,2015Incremental}.
We directly map the optimization objective of the combined reward to the aggregation of various preference models at the policy level, and define the multi-personality generation objective as sampling from a target probability distribution $\pi_\mathrm{MPG}(y|x)$. 
\begin{equation}
    \pi_\mathrm{MPG}(y|x) \propto \mathrm{agg}_{\nabla f}\left(\alpha, \pi_\text{ref}(y|x), \pi_{d}(y|x)\right)
\end{equation}
where $\mathrm{agg}_{\nabla f}$ is an $\nabla f$-function that aggregates each preference model. This target distribution is formally proportional to the aggregation of the characteristics of individual single attribute preference models by the $f$-function.

\subsection{MPG Based on Density Ratio}
\label{3.2}

The core principle guiding our approach stems from a key feature observed in LLM preference alignment training (e.g., DPO~\cite{Rafailov2023DirectPO}): the probability density ratio of an optimized policy model $\pi_\theta(y|x)$ relative to the reference model $\pi_{\text{ref}}(y|x)$ used in its training, $r(y|x) = \frac{\pi_\theta(y|x)}{\pi_{\text{ref}}(y|x)}$, implicitly encodes the preference information learned by the model. This ratio quantifies the model's preference for a particular output sequence $y$ compared to its likelihood under $\pi_{\text{ref}}(y|x)$. We take advantage of this principle to sample from our target multi-personality distribution $\pi_\mathrm{MPG}(y|x)$.

We naturally find that in the rejection sampling algorithm, the core computational load is the density ratio between the target policy and the proposal distribution. And for the density ratio of the target distribution relative to $\pi_{\text{ref}(y|x)}$, it can be derived as follows according to Eq.~\eqref{(5)}:

\begin{equation}
\label{(7)}
\frac{\pi_\mathrm{MPG}(y|x)}{\pi_{\text{ref}}(y|x)} = \frac{1}{Z(x)} \left(\nabla f\right)^{(-1)}\left(\frac{1}{\beta} \sum_{i=1}^{N} \alpha_{i}   \nabla f\left(\frac{\pi_{d_i}(y|x)}{\pi_{\text{ref}}(y|x)}\right)\right)
\end{equation}

Although it is difficult to calculate $Z(x)$, the core idea of MPG is to apply rejection sampling to avoid these complex calculations. The acceptance probability $A(y|x)$ is defined as the ratio between the target distribution and the proposal distribution scaled by an envelope constant $M$:
\begin{equation}
A(y \mid x) = \frac{\pi_{\mathrm{MPG}}(y \mid x)}{M \cdot \pi_{\mathrm{ref}}(y \mid x)}
\end{equation}

Where $M \ge 1$ is chosen such that $M \cdot \pi_{\mathrm{ref}}(y \mid x) \ge \pi_{\mathrm{MPG}}(y \mid x)$ holds for all $y$, ensuring that the acceptance probability always lies within $[0,1]$.
Combine Eq~\eqref{(7)}, for a giver input $x$, $Z(x)$ is a constant that does not change with $y$. Therefore, in the expression above of the target density ratio, $1/Z(x)$, as a multiplicative factor that is the same for all candidates $y$, its effect can be absorbed into the upper bound $M$ of the rejection sampling, the $1/\beta$ is also. And since $(\nabla f)^{(-1)}$ is usually monotonically increasing, it does not change the relative order of its arguments, so that we can get:
\begin{equation}
\frac{\pi_\mathrm{MPG}(y|x)}{\pi_{\text{ref}}(y|x)} \propto\sum_{i=1}^N\alpha_i \nabla f ( \frac{\pi_{d_i}(y|x)}{\pi_{\text{ref}}(y|x)} )
\end{equation}

Defining the individual attribute density ratio as $r_i(y|x) = \frac{\pi_{d_i}(y|x)}{\pi_{\text{ref}}(y|x)}$, representing the preference signal for the $i$-th attribute relative to the reference model. 
In most policy optimizations (the regularization terms are PPO~\cite{Schulman2017ProximalPO} and DPO~\cite{Rafailov2023DirectPO}), the Reverse KL-divergence is used, with $f(x)=x\log x$ and $\nabla f(x)=\log x + 1$. Therefore, we follow this to derive the formula for the density ratio of the target distribution relative to $\pi_{\text{ref}(y|x)}$ as follows:
\begin{equation}
\begin{aligned}
    \frac{\pi_\mathrm{MPG}(y|x)}{\pi_{\text{ref}}(y|x)} &\propto\sum_{i=1}^{N}\alpha_i(\log{r}_i(y|x)+1)\\
    &\propto\sum_{i=1}^{N}\alpha_i{r}_{i}(y|x)
\end{aligned}
\end{equation}

This leads to the acceptance probability:
\begin{equation}
A(y|x)=\frac{\pi_\mathrm{MPG}(y|x)}{M \cdot \pi_{\text{ref}}(y|x)}\propto\frac{\sum_{i=1}^{N}\alpha_i{r}_{i}(y|x)}{M}
\end{equation}
This relationship forms the theoretical cornerstone of our MPG method. It demonstrates that within our rejection sampling framework, the acceptance probability $A(y \mid x)$ for a candidate text sequence is directly governed by the weighted sum of the base density ratios $r_i(y \mid x)$ contributed by the attribute-specific preference models. This efficient representation explicitly encodes multi-dimensional preference information, enabling effective control over personality generation.

In summary, MPG is inspired by preference alignment training and provides a direct, decode-time algorithm for multi-personality generation. Its core advantage lies in significantly reducing the complexity of combinations: Traditional methods may require training a model for each personality combination, resulting in an exponential increase in quantity, while the complexity of MPG is only linearly related to the number of basic attributes.
This principle is crucial for practical applications. Retraining-based algorithms face an impractical scalability challenge when catering to every unique preference. In contrast, MPG is highly scalable: Incorporating a new personality attribute only requires providing its corresponding preference model $\pi_{d_{N+1}}$ and weight $\alpha_{N+1}$, without the need for costly retraining.

\subsection{Speculative Chunk-level based Rejection sampling
(SCR)}
\label{3.3}
We implement MPG in a decoding-time algorithm termed Speculative Chunk-level based Rejection sampling
(SCR). 
Unlike sequence-level rejection sampling, SCR proposes and validates chunks of $k$ tokens at a time, integrating speculative decoding~\cite{Leviathan2022FastIF} with chunk-wise acceptance decisions.
By validating multi-token speculative proposals against the aggregated density ratio, SCR preserves the correctness of rejection sampling while amortizing expensive target-policy validation over longer spans, thereby reducing the frequency of preference model evaluations and enabling efficient parallelization.

\vspace{0.15cm}

\begin{algorithm}[t]
\caption{Speculative Chunk-level based Rejection sampling}
\label{alg:tc_srs}
\begin{algorithmic}[1]
\Require $\pi_{\text{ref}}$, $\{\pi_{d_i}\}_{i=1}^N$, weights $\alpha$, chunk size $k$, window $W$, init bound $\log M_0$, safety margin $\gamma>1$, variance threshold $\tau$
\State $Y \gets \emptyset$, $\mathcal{B} \gets \emptyset$, $\log M \gets \log M_0$, frozen $\gets$ false
\While{not EOS}
    \State \textbf{(Propose)} Sample $C \sim \pi_{\text{ref}}^{(k)}(\cdot|Y)$
    \State Cache $\log\pi_{\text{ref}}(C\mid Y)$
    \ForAll{$i \in \{1,\dots,N\}$ \textbf{in parallel}}
        \State $\overline{\log r}_i(C\mid Y) \gets \log\pi_{d_i}(C\mid Y) - \log\pi_{\text{ref}}(C\mid Y)$
    \EndFor
    \State $S(C\mid Y) \gets \log\!\sum_i \exp\big(\log\alpha_i + \overline{\log r}_i(C\mid Y)\big)$
    \If{$\log u < S(C\mid Y) - \log M$ \textbf{for} $u \sim \mathcal{U}(0,1)$}
        \State $Y \gets Y \Vert C$, push $S(C\mid Y)$ into $                                                                                                                     \mathcal{B}$
    \Else
        \State accepted $\gets$ false
        \For{$j \gets k-1$ \textbf{down to} $1$}
            \State Compute $S(C_{1:j}|Y)$ from cached terms
            \If{$\log u_j < S(C_{1:j}\mid Y) - \log M$}
                \State $Y \gets Y \Vert C_{1:j}$, push $S(C_{1:j}\mid Y)$, accepted $\gets$ true
                \State\textbf{break}
            \EndIf
        \EndFor
        \State\COMMENTLLAMA{\textbf{RS-1 fallback}}
        \If{not accepted} 
        
            \Repeat
                \State Sample $y \sim \pi_{\text{ref}}(\cdot\mid Y)$
                \State Compute $S(y\mid Y)$
            \Until{$\log u_1 < S(y\mid Y) - \log M$}
            \State $Y \gets Y \Vert y$, push $S(y\mid Y)$
        \EndIf
    \EndIf
    \State \textbf{(Update $\log M$)}
    \State\COMMENTLLAMA{\textbf{Warm-up}}
    \If{$|\mathcal{B}| \le W$}
        \State $\log M \gets \log M_0$, frozen $\gets$ false
    \Else
        \If{not frozen}
        \State\COMMENTLLAMA{\textbf{Estimation}}
            \State $\log M \gets \max(\mathcal{B}_W) + \log\gamma$
            \If{$\mathrm{Var}(\mathcal{B}_W) < \tau$} frozen $\gets$ true \EndIf
        \Else
        \State\COMMENTLLAMA{\textbf{Stabilization}}
            \If{$\mathrm{Var}(\mathcal{B}_W) \ge \tau$} frozen $\gets$ false \EndIf
        \EndIf
    \EndIf
\EndWhile
\State \Return{$Y$}
\end{algorithmic}
\end{algorithm}

\noindent\textbf{Notation.}
$Y$: current accepted sequence; 
$C$: proposed $k$-token chunk; 
$C_{1:j}$: prefix of $C$ of length $j$; 
$S(\cdot|\cdot)$: log-score defined in Sec.~\ref{3.3}; 
$\mathcal{B}$: buffer of recent log-scores (size $\le W$); 
$\mathcal{B}_W$: last $W$ log-scores in $\mathcal{B}$;
$\mathrm{Var}(\cdot)$: empirical variance; 
$\gamma$: safety multiplier for $M$; 
$\tau$: variance threshold for freezing $M$ updates;
$\Vert$: sequence concatenation operator.

\paragraph{\textbf{Speculative proposal with rejection-based validation.}}
At decoding step $t$ with current prefix $Y=(y_1,\dots,y_t)$, SCR performs a $k$-step speculative rollout from the reference model:
\begin{equation}
C=(y_{t+1},\dots,y_{t+k}) \;\sim\; \pi_{\text{ref}}^{(k)}(\cdot|Y)
\end{equation}

and validates the whole chunk via a rejection-sampling test against the MPG target policy. 
Let the aggregated ratio be defined as in Sec.~\ref{3.2}:
\begin{equation}
\mathrm{Score}(C\mid Y) \;=\; \sum_{i=1}^N \alpha_i\, r_i(C\mid Y)    
\end{equation}
where $r_i(C\mid Y) = \pi_{d_i}(C\mid Y)/\pi_{\text{ref}}(C\mid Y)$ is the individual attribute density ratio.
The chunk-level acceptance probability is then
\begin{equation}
\label{eq:tc_srs_acc_rate}
A(C\mid Y) \;=\; \min\!\left(1,\;\frac{\mathrm{Score}(C\mid Y)}{M}\right)
\end{equation}
where $M$ is an online upper bound on $\mathrm{Score}(\cdot\mid\cdot)$ (see below). 
For log-domain computations, we also define $S(C\mid Y) = \log \mathrm{Score}(C\mid Y)$.
If a Bernoulli trial with success probability $A(C\mid Y)$ succeeds, the full chunk $C$ is committed; otherwise, a \emph{prefix salvage} cascade is performed:
for $j=k-1,\dots,1$, test the longest prefix $C_{1:j}$ with
$
A(C_{1:j}\mid Y) \;=\; \min\!\left(1,\;\frac{\mathrm{Score}(C_{1:j}\mid Y)}{M}\right),
$
and commit the first accepted prefix (if any). If no prefix is accepted, we fall back to a single-token RS step, preserving correctness while ensuring decoding progress.

\paragraph{\textbf{Numerically-stable computation of $r_i$.}}
To avoid numerical underflow/overflow when $C$ is long, we compute each $r_i(C\mid Y)$ in the log domain:
\begin{equation}
\overline{\log r}_i(C\mid Y) \;=\; \sum_{t'=1}^{|C|} [ \log\pi_{d_i}(y_{t'}|Y_{<t'}) - \log\pi_{\text{ref}}(y_{t'}|Y_{<t'}) ]    
\end{equation}

where $t'$ indexes positions within the chunk $C$ (conditioned on its preceding context $Y_{<t'}$ in the full sequence).
This formulation accumulates per-token log-probability differences, which are already available from the forward passes of the reference and attribute models.
When computing across multiple $i$, we cache $\log\pi_{\text{ref}}$ once and evaluate all $\pi_{d_i}$ in parallel, thus avoiding redundant reference model calls.
The log-score is then obtained stably via
\begin{equation}
S(C\mid Y) = \mathrm{logsumexp}_{i}\!\left(\log \alpha_i \;+\; \overline{\log r}_i(C\mid Y)\right)    
\end{equation}

where $\mathrm{logsumexp}$ uses the standard max-trick. 
The acceptance test in Eq.~\eqref{eq:tc_srs_acc_rate} becomes:
\begin{equation}
u \sim \mathcal{U}(0,1), \quad\text{accept if}\quad \log u \;<\; S(C\mid Y) - \log M    
\end{equation}

where $\mathcal{U}(0,1)$ is a uniform random number in $(0,1)$. This unified log-domain formulation allows the same cached $S(\cdot\mid\cdot)$ to be reused for both full-chunk and prefix salvage tests.

\paragraph{\textbf{Online update of $M$: three-phase sliding-window estimation.}}
We adaptively set $\log M$ from recent log-scores $S(\cdot\mid\cdot)$ in a sliding window of the last $W$ chunks to balance adaptivity and stability:
\begin{enumerate}
    \item \textbf{Warm-up.} For the first $W$ chunks, use a conservative constant $\log M_0$ (or a large prior bound) to avoid premature underestimation.
    \item \textbf{Estimation.} Thereafter, update
    $$
    \log M \leftarrow\max_{t\in\text{window}} S_t + \log\gamma
    $$
    where $S_t$ is the log-score of the $t$-th observed (full or prefix) candidate and $\gamma>1$ is a small safety margin. This corresponds to a sliding-window maximum estimator that tracks the worst-case observed ratio while allowing a buffer against unseen cases.
    \item \textbf{Stabilization.} If $\mathrm{Var}(\{S_t\})<\tau$ within the window, freeze $\log M$ to prevent oscillation; resume updates once the variance exceeds $\tau$.
\end{enumerate}

\paragraph{\textbf{Parallel multi-preference scoring.}}
For each proposed chunk (and its prefixes during salvage), we compute all $\overline{\log r}_i(C\mid Y)$ in parallel:  
(i) cache $\log\pi_{\text{ref}}(C\mid Y)$ once,  
(ii) distribute the $N$ preference models across devices or streams to obtain $\log\pi_{d_i}(C\mid Y)$ concurrently, and  
(iii) aggregate via the shared $S(\cdot\mid\cdot)$ operator.  
Chunking enables batched token- and model-level parallelism, substantially reducing wall-clock latency for large-model validation. Algorithm~\ref{alg:tc_srs} summarizes the full SCR procedure.

\section{Experiments}

To validate the effectiveness of our proposed MPG framework and SCR algorithm, we conduct extensive experiments on two representative multi-personality tasks: MBTI Personality Simulation and Role-Playing. This section outlines our experimental settings and main results, followed by detailed analyses of the method's controllability, inference efficiency, and flexibility.

\subsection{Experimental Settings}

\begin{table*}[ht]
\centering
\renewcommand{\arraystretch}{0.95}
\setlength{\tabcolsep}{5pt}
\caption{Comparison of baseline methods and SCR on the MBTI task. DPO (single) refers to taking the average of $N$ DPO models trained on single dimensions. SCR(ref-Base) and SCR(ref-DPO single) refer to the use of Base and DPO single, respectively, as the reference model to perform the SCR.}
\label{tab:mbti}
\resizebox{0.98\textwidth}{!}{
\begin{tabular}{l ccccc ccccc ccccc c}
\toprule
\multirow{2}{*}{\textbf{Method}} & 
\multicolumn{5}{c}{\textbf{QA}} &
\multicolumn{5}{c}{\textbf{MCQA}} &
\multicolumn{5}{c}{\textbf{16P}} &
\multirow{2}{*}{\textbf{Overall}}\\
\cmidrule(lr){2-6} \cmidrule(lr){7-11} \cmidrule(lr){12-16}& 
{Sty} & 
{Tho} &
{Beh} &
{Nat} & 
{Avg} &
{Sty} & 
{Tho} &
{Beh} &
{Nat} & 
{Avg}  &
{Sty} & 
{Tho} &
{Beh} &
{Nat} & 
{Avg}  
\\
\midrule
\rowcolor[HTML]{E2F0D9}\textbf{Evaluated by GPT-4o} & & & & & & & & & & & & & & & & \\
Base	&3.282 	&3.722 	&3.634 	&3.166 	&3.451 	&2.688 	&3.069 	&3.014 	&2.473 	&2.811 	&2.851 	&3.115 	&3.092 	&2.745 	&2.989 	&3.084 \\

Preference Prompting
	& 3.348	& 3.796	& 3.657	& 3.341	& 3.535	& 2.770	& \underline{3.168}	& 3.131	& 2.538	& 2.902	& 2.888	& 3.273	& 3.287	& 2.854	& 3.076 & 3.171\\
DPO(single)
	&3.559 	&3.804 	&3.885 	&4.003 	&3.813 	&1.911 	&2.135 	&2.053 	&2.079 	&2.044 	&3.149 	&3.543 	&\underline{3.442} 	&3.809 	&3.486 	&3.114 \\

DPO Soups
& 3.458 & 3.937 & 3.850 & 3.494 & 3.685 & 2.751 & 3.161 & \underline{3.196} & 2.557 & \underline{2.916} & 2.936 & 3.203 & 3.345 & 2.892 & 3.094 & 3.232 \\
MOD& 3.517 & \underline{4.049} & \underline{3.911} & \underline{4.066} & \underline{3.886} & 1.727 & 1.846 & 1.758 & 1.946 & 1.819 & \underline{3.272} & \underline{3.692} & 3.431 & \underline{3.864} & \underline{3.565} & 3.090 \\
SCR(ref-Base)
& \underline{3.583} & 4.030 & 3.878 & 3.677 & 3.792 & \textbf{2.832} & \textbf{3.305} & \textbf{3.402} & \underline{2.692} & \textbf{3.058} & 2.933 & 3.317 & 3.395 & 3.008 & 3.163 & \underline{3.338} \\
SCR(ref-DPO single)
& \textbf{3.927} & \textbf{4.322} & \textbf{4.444} & \textbf{4.271} & \textbf{4.241} & \underline{2.789} & 3.026 & 2.866 & \textbf{2.803} & 2.846 & \textbf{3.481} & \textbf{3.900} & \textbf{3.786} & \textbf{4.000} & \textbf{3.792} & \textbf{3.626} \\

\midrule
\rowcolor[HTML]{E2F0D9}\textbf{Evaluated by DeepSeek-R1} & & & & & & & & & & & & & & & & \\		
Base	&3.375 	&4.025 	&3.978 	&3.404 	&3.696 	&2.617 	&3.523 	&3.602 	&2.598 	&3.085 	&2.952 	&3.546 	&3.759 	&3.035 	&3.323 	&3.368 \\

Preference Prompting
& 3.438 & 4.093 & 4.139 & 3.508 & 3.794 & 2.785 & 3.710 & 3.777 & 2.697 & 3.242 & 3.058 & 3.706 & 3.831 & 3.172 & 3.442 & 3.493 \\
DPO(single)
	&3.792 	&3.989 	&4.113 	&4.263 	&4.039 	&2.016 	&2.160 	&2.185 	&2.115 	&2.119 	&3.570 	&3.771 	&3.921 	&4.346 	&3.902 	&3.353 \\

DPO Soups
& 3.569 & \underline{4.296} & 4.335 & 3.768 & 3.992 & 2.821 & \underline{3.810} & \textbf{3.968} & 2.848 & \underline{3.362} & 3.117 & \underline{3.911} & \underline{4.119} & 3.217 & 3.591 & 3.648 \\
MOD& \underline{3.825} & 4.077 & 4.207 & \underline{4.399} & \underline{4.127} & 1.746 & 1.798 & 1.830 & 1.878 & 1.813 & \underline{3.691} & 3.892 & 3.941 & \underline{4.355} & \underline{3.970} & 3.303 \\
SCR(ref-Base)
& 3.656 & 4.286 & \underline{4.337} & 3.833 & 4.057 & \underline{2.876} & \textbf{3.824} & \underline{3.871} & \underline{2.912} & \textbf{3.373} & 3.147 & \underline{3.911} & 3.992 & 3.314 & 3.591 & \underline{3.674} \\
SCR(ref-DPO single)
& \textbf{4.153} & \textbf{4.401} & \textbf{4.713} & \textbf{4.584} & \textbf{4.463} & \textbf{2.920} & 3.501 & 3.244 & \textbf{3.084} & 3.137 & \textbf{3.900} & \textbf{4.153} & \textbf{4.147} & \textbf{4.422} & \textbf{4.156} & \textbf{3.919} \\

\bottomrule
\end{tabular}}
\vskip -1em
\end{table*}
\begin{table*}[ht]
\centering
\renewcommand{\arraystretch}{0.95}
\setlength{\tabcolsep}{5pt}
\caption{Comparison of baseline methods and SCR on Role-Playing task. DPO (single) refers to taking the average of $N$ DPO models trained on single dimensions. SCR(ref-Base) and SCR(ref-DPO single) refer to the use of Base and DPO single, respectively, as the reference model to perform the SCR. }\label{tab:role}
\resizebox{0.98\textwidth}{!}{
\begin{tabular}{l cccc cccc cccc}
\toprule
\multirow{2}{*}{\textbf{Method}} &
\multicolumn{4}{c}{\textbf{Evaluated by GPT-4o}} &
\multicolumn{4}{c}{\textbf{Evaluated by DeepSeek-R1}} &
\multicolumn{4}{c}{\textbf{Reference-based Evaluation}}\\
\cmidrule(lr){2-5} \cmidrule(lr){6-9} \cmidrule(lr){10-13}&
PR & 
RM &
Hl &
Avg &
PR & 
RM &
Hl &
Avg &
BLEU &
ROGUE-1 &
BERTScore &
PPL
\\
\midrule
Base	&3.580 	&3.770 	&3.471 	&3.607 &
3.418 	&3.778 	&4.098 	&3.765 &
0.010 	&0.088 	&0.762 	&43.984 \\

Preference Prompting
 & 3.720  & 3.853  & 3.832  & 3.802 & 3.582  & 3.821  & 4.283  & 3.895 &  0.024 	&0.127 	&0.823 	&\underline{41.860} 
\\
DPO(single)
 & 3.823 	&4.240 	&4.137 	&4.066 &
3.744 	&4.106 	&4.563 	&4.137 &
0.058 	&0.167 	&0.868 	&48.262 \\
DPO Soups
 & 3.818  & 3.998  & 3.871  & 3.896 & 3.692  & 3.857  & 4.379  & 3.976 & 0.022 	&0.130 	&0.823 	&43.594 
\\
MOD & \underline{4.020}  & \textbf{4.298}  & \underline{4.170}  & \underline{4.166} & \underline{3.970} & \textbf{4.187}  & \underline{4.586}  & \textbf{4.241} & \underline{0.082} 	&\underline{0.187} &	\underline{0.870} 	&47.188 
\\
SCR(ref-Base)
 & 3.874  & 3.997  & 3.861  & 3.911 & 3.722  & 3.833  & 4.414  & 3.990  & 0.034 	&0.133 	&0.829 	&\textbf{36.694} 
\\
SCR(ref-DPO single)
 & \textbf{4.030}  & \underline{4.290}  & \textbf{4.192}  & \textbf{4.167} & \textbf{3.976}  & \underline{4.166}  & \textbf{4.596}  & \underline{4.230} &  \textbf{0.085} 	&\textbf{0.188} 	&\textbf{0.897} 	&43.335 
\\

\bottomrule
\end{tabular}
}
\end{table*}

We evaluate SCR on two representative and valuable multi-personality generation tasks. All single-attribute DPO models were trained based on the Llama-3-8B-Instruct model~\footnote{\url{https://huggingface.co/meta-llama/Meta-Llama-3-8B-Instruct}} using LoRA. The LoRA configuration was set with $r=8$, $\alpha=16$, and dropout rate $0.05$. Training utilized the AdamW optimizer with a learning rate of $5 \times 10^{-5}$, weight decay of $0.01$, and a batch size of 16. Training was performed for 3 epochs. Gradient accumulation was used for 8 steps. We employed a cosine learning rate scheduler with warm-up for 100 steps. The DPO loss temperature $\beta_{DPO}$ was set to 0.1. 

For our configured SCR algorithm with the following parameters. The speculative chunk size $k$ was set to 4. For the dynamic update of the rejection bound $M$, we employed a sliding-window estimation strategy with a window size of $W=20$. A conservative initial bound $\log M_0$ was used during the warm-up phase. The safety margin $\gamma$ was set to 1.2 to ensure the bound remains robust against unseen high-scoring candidates. The variance threshold for freezing M updates, $\tau$, was set to 0.01 to prevent oscillation. The reference model $\pi_\text{ref}$ generated candidate chunks using a sampling temperature of 0.7 and top-p of 0.9 to ensure diversity in proposals.

\subsubsection{MBTI Personality Simulation}
\label{sec:4.1.1}
\paragraph{Datasets} For the MBTI Personality Simulation task, on the training stage, we construct a specialized training dataset derived from \texttt{pandalla/Machine\_Mindset\_MBTI\_dataset}~\footnote{\url{https://huggingface.co/datasets/pandalla/Machine_Mindset_MBTI_dataset}} that captures the four fundamental MBTI dimensions $D_\text{MBTI}$ which contains data pairs $(y_w,y_l)_i$, where $y_w$ is preferred over $y_l$ on the $i$-th dimension. For evaluation, we employ three curated benchmarks: 1) MBTI-QA~\footnote{\url{https://huggingface.co/datasets/pandalla/Machine_Mindset_MBTI_dataset}} for instruction-following question answering, 2) MBTI-MCQA~\cite{Pan2023DoLP} for multiple-choice answering, and 3) MBTI-16P~\footnote{\url{https://www.16personalities.com/free-personality-test}} containing items from the 16Personalities psychometric instrument. These datasets systematically assess personality-specific response patterns through carefully constructed diagnostic questions.

\paragraph{Evaluation Metrics} We primarily employ the LLM-as-a-Judge while using GPT-4o and DeepSeek-R1 as the main evaluators. We designed detailed evaluation prompts and scoring rubrics for each task. The evaluation dimensions assessed by the LLM judges include: 

\textbf{1) Style (Sty):} Evaluates the alignment of the answer's language style, including word, sentence, and tone, with the target personality.

\textbf{2) Thought (Tho):} Assesses whether the answer reflects the target personality's characteristic thinking patterns, such as their logic, values, and priorities in decision-making.

\textbf{3) Behavior (Beh):} Measures the extent to which the answer demonstrates behavioral traits specific to the personality, such as being proactive, organized, or empathetic in expression.

\textbf{4) Naturalness (Nat):} Judges the overall quality of the language, including its fluency and conciseness, and ensures the response avoids sounding robotic or like a forced imitation.

\subsubsection{Role-Playing}
\label{sec:4.1.2}
\paragraph{Datasets}  For the Simple Role-Playing task, we constructed the dataset by adapting the methodology from \texttt{ALOE}~\footnote{\url{https://github.com/ShujinWu-0814/ALOE}}, focusing on user profile and target personality. The training set includes annotations generated by the ChatGPT-4o model~\cite{Hurst2024GPT4oSC}. The evaluation set, a subset of this dataset, contains conversation contexts grounded in user profile descriptions and specific persona. Generated responses were evaluated based on their ability to effectively simulate the described profile and persona within these contexts.
\paragraph{Evaluation Metrics} We adopt the same evaluation methodology as for the MBTI personality simulation task, using LLM-as-a-Judge~\cite{Zheng2023JudgingLW}. The evaluation dimensions are as follows:

\textbf{1) Profile Relevance (PR):} Assesses the alignment of the generated content with the user's specified identity, background details, and interests as described in their profile.

\textbf{2) Persona Match (PM):} Evaluates the answer's consistency with the provided personality traits, including the appropriate tone of voice, emotional expression, and behavioral patterns.

\textbf{3) Humanlikeness (Hl):} Judges the overall quality of the language for its naturalness, concision, and emotional authenticity, ensuring the response is fluent and avoids artificial phrasing.

Additionally, we also calculated scores such as \textbf{BLEU}, \textbf{ROUGE-1 F1}, \textbf{BERTScore}, and \textbf{Perplexity} to provide linguistic and semantic evaluation dimensions of the quality of the content.

\subsection{Baselines}  
We compare SCR with the baselines, which combine the capabilities of existing models at decode time without requiring difficult-to-obtain external models.  We aim to compare our approach within the scope of methods achievable using readily available resources.

\textbf{Preference Prompting (PP)}~\cite{Jang2023PersonalizedSP} implements personality conditioning through explicit attribute descriptions in prompt engineering, applied across base model, single-preference model, and specialized model(use in  Sec.~\ref{4.6}) configurations. \textbf{DPO Soups (a variant of Personalized Soups)}~\cite{Jang2023PersonalizedSP,Zhou2023BeyondOA} acquires the LoRA parameters of independently - trained single - preference DPO models. Following weighted averaging, these parameters are applied to $\pi_{\text{ref}}$ to generate text with multi-personality. \textbf{MOD}~\cite{Shi2024DecodingTimeLM} performs a weighted linear combination of the output logits from single - attribute DPO models during decoding and samples based on the combined logits. \textbf{SCR}: Our Speculative Chunk-level based Rejection sampling with parallel multi-preference scoring.

\begin{figure*}[t]
\centering
  \includegraphics[width=\linewidth]{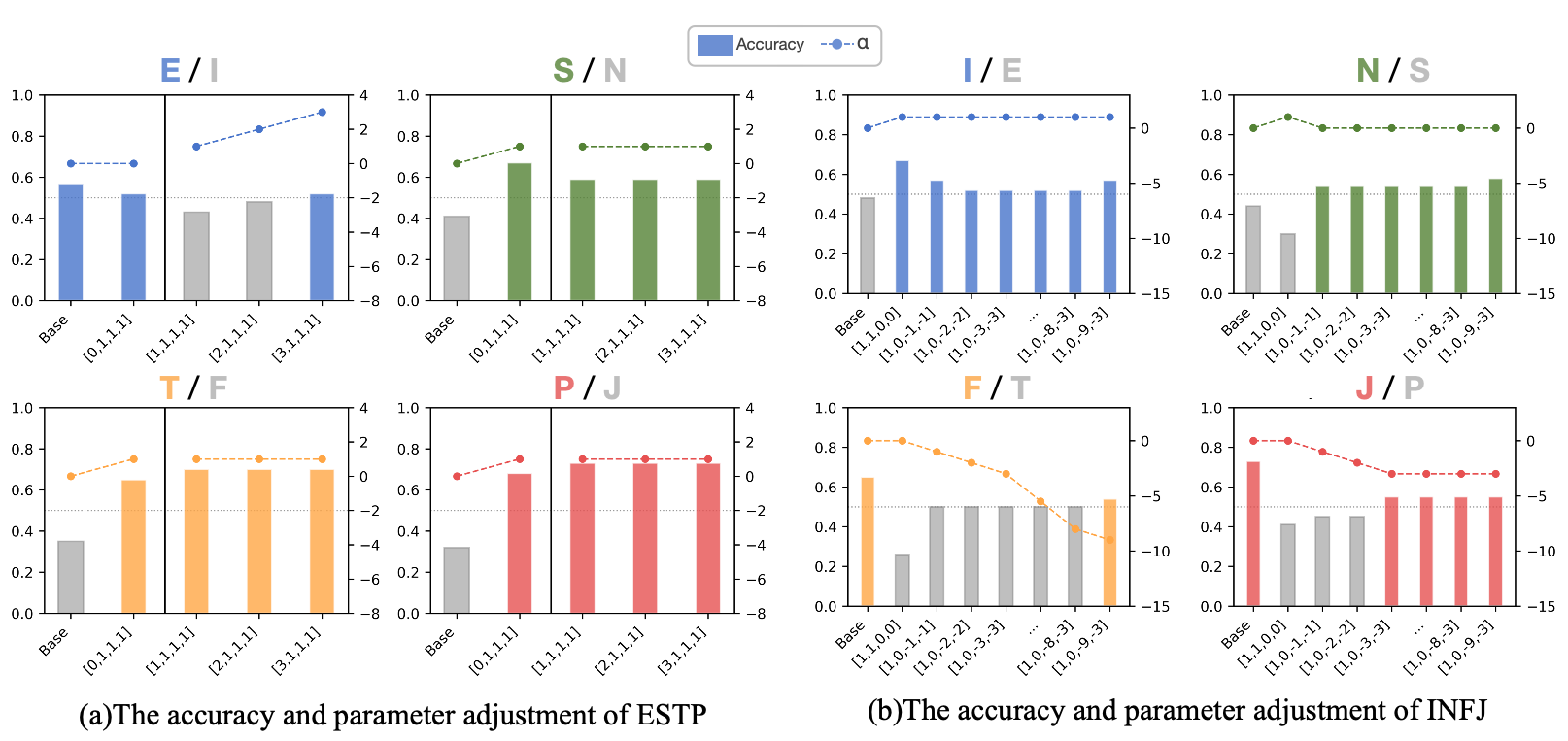} 
  \caption {Iterative tuning process for the $\alpha$. Bars indicate prediction accuracy (left axis) for each MBTI dimension; dashed lines track $\alpha$ values (right axis) at each optimization step. (a) ESTP-targeted tuning shows monotonic $\alpha$ progression. (b) INFJ-targeted tuning demonstrates non-monotonic adjustments with negative $\alpha$ phases.}
  \label{fig:alpha_tuning}
\end{figure*}

\subsection{Main Results}

Table~\ref{tab:mbti} presents the performance of various methods on the MBTI personality simulation task. The reported scores represent the average performance across three representative target personality types: ESTP, INFJ, and ENTJ. Considering the Overall score, SCR(ref-DPO single) and SCR(ref-Base), achieve the highest scores for both evaluation on GPT-4o and DeepSeek-R1,and DPO single refers to a model that conducts DPO training only on a single dimension. Compared to the Base model, SCR(ref-DPO single) demonstrates a substantial improvement of 16.36\%-17.57\%, while SCR(ref-Base) shows an improvement of 8.24\%-9.09\%.

Notably, on the QA and 16P datasets, SCR(ref-DPO single) and MOD ranked among the top two performers. However, on the MCQA dataset, both MOD and DPO(single) methods performed poorly, while SCR(ref-Base) achieved the highest score. Since the MOD method relies on linearly combining the prediction logits of DPO(single) models, its performance is directly impacted by their limitations on such data distributions. In contrast, our SCR method leverages sampling from a robust reference model ($\pi_{\text{ref}}$) (either Base or DPO single model) and combines preference signals through density ratios. The use of a strong reference model in SCR provides a more stable foundation for generation, especially when individual preference models ($\pi_{d_i}$) might be brittle on specific data distributions. This inherent robustness, particularly when using the Base model as $\pi_{\text{ref}}$, explains SCR(ref-Base)'s superior performance on the MCQA dataset and its strong Overall ranking.
In addition, the consistent superiority of SCR across heterogeneous MBTI target types suggests that the method can adaptively balance personality dimensions, reflecting a stronger generalization capacity than other baselines.

Other baselines, such as Preference Prompting and DPO Soups, generally improve upon the Base model but are consistently surpassed by our SCR methods. This performance gap remains stable across different evaluation backbones, highlighting that the advantage of SCR is not tied to a specific evaluation model but stems from its decoding-time aggregation mechanism.

Table~\ref{tab:role} summarizes the performance on Role-Playing task. SCR (ref- DPO single) and MOD are better than baselines such as PP and DPO Soups, achieving comparable scores on both evaluation metrics, usually the highest. While MOD demonstrates slightly higher scores on some LLM Evaluation, SCR(ref-DPO single) often leads in most LLM Evaluation metrics and in terms of reference similarity metrics. This may reflect the different combination mechanisms: the logit combination of MOD may often affect the representation of properties at the single token level, while the density-ratio based SCR have an advantage in generating text that is more globally coherent and more similar to the reference distribution while achieving good results.

\subsection{Alpha Analysis}

In the MBTI task, we performed an iterative adjustment of the $\alpha$ = [\sethlcolor{EI}\hl{$\alpha_E$}, \sethlcolor{SN}\hl{$\alpha_S$}, \sethlcolor{TF}\hl{$\alpha_T$}, \sethlcolor{PJ}\hl{$\alpha_J$}]. This process was guided by the model's prediction accuracy on each MBTI dimension as measured on the MCQA dataset. We aimed to find the optimal $\alpha$ that allowed the model to converge towards the target personality.

Figure~\ref{fig:alpha_tuning}(a) illustrates the tuning process for the ESTP target personality. Starting from baselines including the Base model performance and an initial $\alpha$ configuration (e.g., [1,1,1,1]), we iteratively adjusted $\alpha$ values, primarily increasing the weights corresponding to the ESTP dimensions (E, S, T, P) where performance required improvement. As shown, this process generally led to increased accuracy across these relevant dimensions, enabling alignment with the target profile.

Tuning for the INFJ target personality is more complex. As shown in Figure~\ref{fig:alpha_tuning}(b), initial $\alpha$ adjustments resulted in decreased accuracy on the N, F, and J dimensions. To address these potential conflicts, our tuning strategy allows negative values for certain dimensions to balance the overall effect. In our chunk-level scoring formulation (Sec.~\ref{3.3}), the aggregated score for a candidate chunk $C$ is computed as:
\begin{equation}
Score(C|Y; \alpha) = \sum_{i=1}^N \alpha_i , r_i(C|Y)
\end{equation}
Negative values of $\alpha_i$ are allowed to capture complex interdependencies and conflicts between preference dimensions, and to suppress specific traits.
For example, in the MBTI personality simulation, 'E' (Extraversion) and 'T' (Thinking) preferences may occasionally conflict; assigning a negative weight to 'E' can help suppress overly outgoing expressions in contexts where 'T' is dominant.

Since $S(C|Y;\alpha)$ may become negative when some $\alpha_i < 0$, we ensure the correctness of the rejection sampling process by defining the unnormalized reward:
\begin{equation}
R(C|Y) = \max\left\{ 0, \ \text{Score}(C|Y; \alpha) \right\}
\end{equation}
and the chunk-level acceptance probability as:
\begin{equation}
A(C|Y) = \frac{R(C|Y)}{M}
\end{equation}
where $M$ is the sliding-window estimate of the maximum reward over recent chunks. This guarantees $A(C|Y) \ge 0$ at all times.
If $Score(C|Y; \alpha) < 0$, then $R(C|Y) = 0$ and $A(C|Y) = 0$, meaning the chunk is always rejected, effectively eliminating candidates strongly inconsistent with the desired composite preference.

\begin{figure}[h]
\centering
  \includegraphics[width=0.95\linewidth]{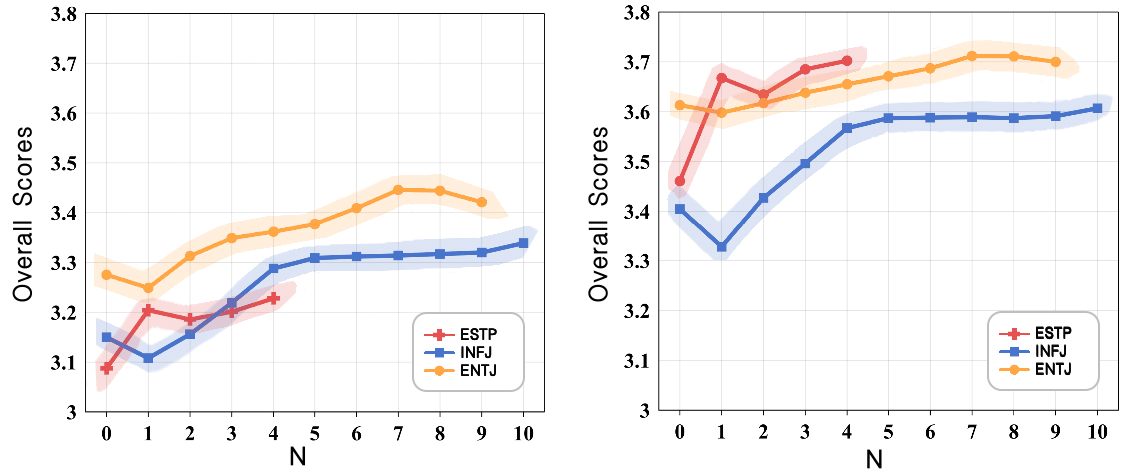} 
  \caption {The variation of model Overall Scores with the number of adjustments of $\alpha$.}
  \label{fig.adjust}
\end{figure}

This design allows $\alpha_i$ to be flexibly tuned (even to negative values) without breaking the probability semantics of rejection sampling, while providing a principled way to modulate or suppress certain personality traits during generation. Figure~\ref{fig.adjust} illustrates how the overall model score varies with successive adjustments of $\alpha$. And the optimal $\alpha$ combination derived from this process for INFJ was found to be $[1, 0, -9, -3]$.

Table~\ref{tab:mbti} reports the performance of SCR(ref-Base) on MBTI task using the optimal $\alpha$ combination determined for each specific target personality.
It is noteworthy that even with un-tuned, uniform α weights, our method demonstrates a significant improvement over the base model. The results presented for the Role-Playing task in Table~\ref{tab:role}, for instance, utilize these un-tuned parameters, showcasing the robust performance of SCR.

\subsection{Efficiency Analysis}
\label{subsec:efficiency}
We evaluate the efficiency of our proposed SCR in the Role-Playing setting ($N=2$ preference models, chunk size $k=4$) against four baselines: \textbf{Base} model, \textbf{DPO Soups}, \textbf{MOD} , \textbf{Seq-RS}(Sequence-level Rejection Sampling), and \textbf{Token-RS} (Token-level Rejection Sampling). 
All experiments are conducted to generate responses for 100 prompts with a maximum of 128 new tokens. 
Each method is run three times under identical hardware/software conditions, and we report the averaged results. 

We measure: \textbf{Throughput} (tokens/sec, $\uparrow$ better), \textbf{Latency} (seconds/sequence, $\downarrow$ better), \textbf{Forward Pass Overhead} (average number of model forward passes per token, $\downarrow$ better), and \textbf{Rejection Rate} ($\%$, $\downarrow$ better).Throughput and latency are measured end-to-end; forward pass counts reflect the number of full forward evaluations of large models (reference + preference models) during decoding.
\begin{table}[h]
\centering
\small
\caption{Efficiency comparison of different methods for Role-Playing task ($N=2$, chunk size $k=4$).$\uparrow$ indicates higher is better, while $\downarrow$ indicates lower is better.}
\label{tab:efficiency_n2}
\resizebox{0.48\textwidth}{!}{
\begin{tabular}{lcccc}
\toprule
\multirow{2}{*}{\textbf{Method}} & \textbf{Throughput  $\uparrow$} & \textbf{Latency  $\downarrow$} & \textbf{FwdPass $\downarrow$} & \textbf{Reject Rate  $\downarrow$} \\
 & (tok/s) & (s/seq) & (per tok) & (\%) \\
\midrule
Base       & 120.0 & 1.07 & 1.0 & --- \\
DPO Soups                         & 118.0 & 1.08 & $\sim$1.0 & --- \\
MOD      & 60.0  & 2.13 & 2.0 & --- \\
Seq-RS      & 11.0  & 11.64 & 10.0 & 60.0 \\
Token-RS                          & 30.0  & 4.27  & 4.62 & 25.0 \\
\midrule
SCR      & 97.0 & 1.20 & 1.4  & 30.0 \\
\bottomrule
\end{tabular}
}
\end{table}

\noindent
From Table~\ref{tab:efficiency_n2}, we observe: single-model methods (Base, DPO Soups) achieve the highest throughput and lowest latency as they require only one forward pass per token; MOD incurs linear overhead in $N$ due to per-model scoring at each step; sequence-level RS suffers from prohibitively high rejection rates, resulting in extremely low throughput; token-level RS improves efficiency but still requires multiple model evaluations per token; SCR achieves a near single-model latency (1.20s/seq) while dynamically fusing multi-preference models, thanks to chunk-level amortization and parallel scoring.

We also conducted the same efficiency analysis in the MBTI task ($N=4$). 
Compared to $N=2$, all multi-model methods show higher forward-pass overhead and rejection rates, with Seq-RS and Token-RS suffering the most. 
However, SCR maintains its efficiency advantage, with throughput still $2\times$--$10\times$ higher than other multi-model baselines.
Overall, SCR offers a practical trade-off between flexibility and speed.

\subsection{Leveraging Specialized Reference Models}
\label{4.6}

Our theoretical framework, grounded in the density ratio principle, is applicable to preference models $\pi_{d_i}$ trained under specific $f$-divergences, such as the reverse KL-divergence common in DPO and PPO. Within this framework, the choice of the reference model $\pi_{\text{ref}}$, which serves as the proposal distribution in our sampling algorithm, offers significant flexibility. A stronger, more domain-aligned $\pi_{\text{ref}}$ can potentially provide higher-quality initial candidates, which can then be steered by our preference models.

To investigate this, we evaluated the performance of SCR when using a different, more specialized reference model. Specifically, we replaced the default base model with a strong, open-source model pre-trained for Role-Playing~\cite{Cui2023MachineMA,Wang2025CoSERCL}\footnote{\url{https://modelscope.cn/organization/FarReelAILab}}\textsuperscript{,}\footnote{\url{https://huggingface.co/Neph0s/CoSER-Llama-3.1-8B}}, while keeping our DPO-trained preference models $\pi_{d_i}$ unchanged.

As shown in Table~\ref{tab:ext}, configuring SCR with this specialized model as the reference (SCR(ref-specialized)) outperforms the specialized model used alone (Base(specialized model)) on both tasks, achieving higher Overall scores. 
This outcome demonstrates a key advantage of our framework: SCR can effectively leverage the strong generative prior of a specialized model as its proposal distribution and successfully steer its outputs towards the desired multi-personality target using the preference signals from the DPO models. This highlights the method's capability to act as a flexible, decoding-time alignment layer on top of existing powerful model assets, achieving performance gains without costly retraining or adaptation of the specialized model itself.

\begin{table}[ht]

\centering

\caption{Performance of SCR using a Specialized Model as the reference. }\label{tab:ext}
\resizebox{0.4\textwidth}{!}{
\tiny
\begin{tabular}{l l c}
\toprule
\textbf{Method} & & \textbf{Overall} \\
\toprule

\multirow{2}{*}{\textbf{Base(specialized model)}} &MBTI	& 3.871 \\ 
&Role-Play	& 4.257 \\
\midrule

\multirow{2}{*}{\textbf{SCR(ref-specialized)}}&MBTI	& 3.947 \\
&Role-Play	& 4.317 \\

\bottomrule
\end{tabular}}

\end{table}

\section{Conclusion}
We introduce MPG with Speculative Chunk-level based Rejection sampling (SCR), a flexible and robust decoding-time framework for multi-preference LLM generation. 
Experiments on MBTI and Role-Playing tasks demonstrate significant gains in controllability and personalization while maintaining practical efficiency, with SCR achieving a favorable trade-off between quality and computational cost compared to strong baselines. 
Future work will explore richer preference combination functions and further optimization of chunk-level acceptance strategies, advancing controllable and efficient personalized generation.

\section*{Ethical Impact Statement}

Our proposed MPG framework operates entirely at the decoding-time, composing pre-existing, independently trained preference models without requiring any new data collection or model retraining. Consequently, our method does not introduce novel ethical risks beyond those inherent in the base and preference models used. We advocate for the responsible use of MPG in benign applications such as enhancing personalized assistants and creative simulations.

\section*{Acknowledgment}
This work was supported by the Strategic Priority Research Program of the CAS (No. XDB0680302),
and the National Natural Science Foundation of China (No.U21B2046, No.62202448).


\bibliographystyle{ACM-Reference-Format}
\bibliography{sample-base}

\appendix
\begin{table*}[!t]
    \small
    \caption{Evaluation prompt for MBTI task}
    \label{tab:prompt_e_mbti}
    {\ttfamily
    \begin{tabularx}{\linewidth}{X}
    \toprule
    You are an expert in the assessment of MBTI personality language style and thought congruence, please evaluate the following conversation. Your task is to determine whether the responses are consistent with the given MBTI personality traits based on the dimensions of language style and core ideas.\\
    Please pay particular attention to: \\
     - Do not give high scores for responses that include direct references to MBTI types or related terminology (e.g., "I am an INFP"); \\
     - Higher scores should be given to responses that naturally reflect the target personality traits; \\
     - You will be judging the consistency of the styles and ideas based on the following personality descriptions.\\
     The given MBTI description and performance is as follows: \{MBTI\_description\}.
    The dialog is as follows: \\
    \text{\text{[}Question\text{]}}: \{Question\} \\
    \text{\text{[}Answer\text{]}}: \{Answer\} \\
    The scoring dimensions are as follows (out of 5 points for each dimension): \\
    1. **Linguistic Style Match **: Whether or not the expression is in line with the personality's linguistic style (e.g., enthusiastic, calm, etc.).\\
    2. **Thinking Logic Consistency **: Whether or not it reflects the personality's typical way of thinking (e.g., based on facts or values, focusing on logic or feelings).\\
    3. **Behavior Consistency**: Whether or not it demonstrates the personality's typical behavioral motivations (e.g., planning or flexibility, preference for social interaction or solitude, etc.).\\
    4. **Overall naturalness**: Whether the answer is concise, natural, coherent, and not intentionally imitative.\\
    Please output a JSON format score (1 to 5 points per item), including the overall score and a short description. For example: \\
    \{\{ \\
    "style": 4, \\
    "thought": 5, \\
    "behavior": 5, \\
    "naturalness": 4.5, \\
    "reason": "The response is more linguistically consistent with the style described, the logic of thought and motivation for behavior match, and the language is natural but shows slight signs of deliberate parody."\\
    \}\}\\
    \bottomrule
    \end{tabularx}
    }

\end{table*}

\begin{table*}[!t]
    \small
    \caption{Evaluation prompt for Role-Playing task}
    \label{tab:prompt_e_role}
    {\ttfamily
    \begin{tabularx}{\linewidth}{X}
    \toprule
    You are an expert in evaluating AI-generated role-playing conversations. Your task is to assess whether the AI successfully role-plays a human user based on the given **user profile** and **personality traits**, and interacts naturally with another intelligent assistant.\\

    You will be given the following:\\
    \text{[User Profile]}: \{profile\}\\
    \text{[Personality Traits]}: \{personalities\}\\
    \text{[Roleplayed Message]}: \{message\}\\
    
    Please rigorously evaluate the message on the following criteria (each scored 1 to 5):\\
    1. **Relevance to User Profile**: Does the content deeply reflect the user’s identity, background, or interests, beyond surface-level mention?\\
    2. **Personality Consistency**: Does the communication style of the message match the provided personality traits (e.g., introverted vs. extroverted)? \\
    3. **Human-likeness and Appeal**: Is the response natural, concise, emotionally engaging, and free from robotic or overly generic expressions? Deduct points for robotic phrasing, vague expressions, repetitive templates, or signs of “AI-ness”.\\
    
    Please output a JSON format score (1 to 5 points per item), including the overall score and a short description. For example: \\
    \{\{\\
    "profile\_relevance": 3.0,\\
    "personality\_match": 2.5,\\
    "humanlikeness": 3.5,\\
    "reason": "The message touches on the user's interests but lacks depth. The tone is inconsistent with the described personality and feels somewhat templated."\\
    \}\}\\

    \bottomrule
    \end{tabularx}
    }

\end{table*}
\begin{table*}[!t]
    \small
    \caption{Prompt for Preference Prompting on MBTI task}
    \label{tab:prompt_p_mbti}
    {\ttfamily
    \begin{tabularx}{\linewidth}{X}
    \toprule
    "E": "You are an 'E' (Extraversion) person in MBTI's personality, good at interacting with others, and your energy comes from interacting with others. Specific manifestations: energetic when interacting with others, enthusiastic and proactive, willing to express, taking action before thinking, and quick to react. ",\\
    "I": "You are an 'I' (Introversion) person in MBTI's personality, good at independent thinking, and your energy comes from self-reflection. Specific manifestations: quiet and introverted, energetic when alone, thinking before action, and thoughtful and thoughtful. ",\\
    "S": "You are an 'S' (Sensing) person in MBTI's personality, tending to focus on specific details of the real world and relying on your five senses to concentrate on the present moment. Specific manifestations: lack of interest in empathy and theory, traditional approach, preference for using known skills, preference for detailed descriptions, and emphasis on depth rather than breadth. ",\\
    "N": "You are an 'N' (iNtuition) person in MBTI's personality, tending to focus on abstract concepts and future possibilities, relying on your sixth sense to pursue novelty. Specific manifestations: interest in concepts and theories, creativity, emphasis on possibilities, liking to learn new technologies, holistic thinking, and valuing breadth over depth.",\\
    "T": "You are a 'T' (Thinking) person in MBTI's personality, relying mainly on logic and analysis when making decisions, pursuing objectivity and rationality. Specific manifestations: reasoning, questioning, treating everyone equally, being frank, recognizing emotions that are logical, being good at discovering shortcomings, and tending to criticize.",\\
    "F": "You are an 'F' (Feeling) person in MBTI's personality, relying mainly on emotions and values when making decisions, and valuing emotions and interpersonal relationships. Specific manifestations: forward thinking, empathetic thinking, emphasis on exceptions to rules, compassion, tenderness, emotionalism, lack of logic, and emphasis on maintaining network resources. ",\\
    "J": "You are a 'J' (Judging) person in MBTI's personality, who likes to be planned and organized, pursuing clear goals and organization. Specific manifestations: Joyful decision-making, prioritizing work, setting goals, completing tasks on time, focusing on results and schedule management, and emphasizing efficiency. ",\\
    "P": "You are a 'P' (Perceiving) person in MBTI's personality, who likes flexibility and openness, pursues possibilities and change. Specific manifestations: being happy when there are multiple choices, enjoying first before working, frequently changing goals, relaxing casually, and paying attention to the process. ",\\

    \bottomrule
    \end{tabularx}
    }

\end{table*}

\begin{table*}[!t]
    \small
    \caption{Prompt for Preference Prompting on Role-Playing task}
    \label{tab:prompt_p_role}
    {\ttfamily
    \begin{tabularx}{\linewidth}{X}
    \toprule
    Your task is to play the role of a person with the following profile and personalities traits and chat with a chatbot:\\
    Profile: \{profile\}\\
    Personalities: \{personality\}\\
    Please ignore the gender pronouns in the personalities and use the correct pronouns based on the given profile.\\
    Please follow the requirements:\\
    1. You should determine the topic of conversation based on the given profile. You should determine the conversational styles based on the given personalities.\\
    2. Keep in mind that you are chatting with a friend instead of a robot or assistant. So do not always seek for advice or recommendations.\\
    3. Do not include any analysis about how you role-play this user. Only output your messages content.\\
    Now, initiate the conversation with the chatbot in whatever way you like. Please always be concise in your questions and responses and remember that you are pretending to be a human now, so you should generate human-like language. \\
    \bottomrule
    \end{tabularx}
    }

\end{table*}

\end{document}